\documentclass{article}

\usepackage{PRIMEarxiv}
\usepackage{graphicx} 
\usepackage{caption}
\usepackage{booktabs}       
\usepackage{amsfonts}       
\usepackage{nicefrac}       
\usepackage{microtype}      
\usepackage{subfigure}
\usepackage{graphicx}
\usepackage[utf8]{inputenc}
\usepackage{balance}
\usepackage{fontawesome}
\usepackage{multirow,array}
\usepackage{tikz}
\usepackage{pgfplots}
\usetikzlibrary{pgfplots.dateplot}
\usepackage{pgfplotstable}

\usepackage{filecontents}
\usepgfplotslibrary{fillbetween}
\usetikzlibrary{patterns}
\usetikzlibrary{backgrounds}
\usetikzlibrary{arrows,automata}
\graphicspath{ {images/} }
\usepackage{wrapfig}
\usepackage{diagbox}
\usepackage{adjustbox}
\usepackage{tcolorbox}
\usepackage{algorithm}
\usepackage{algorithmic}
\usepackage{multicol}
\usepackage[utf8]{inputenc} 
\usepackage[T1]{fontenc}    
\usepackage{hyperref}       
\usepackage{url}            
\usepackage{booktabs}       
\usepackage{amsfonts}       
\usepackage{nicefrac}       
\usepackage{microtype}      
\usepackage{lipsum}
\usepackage{fancyhdr}       
\usepackage{graphicx}       
\graphicspath{{media/}}     

\title{A Mental Model Based Framework of Trust}
\author{
  Zahra Zahedi \\
  SCAI, ASU \\
  \texttt{zzahedi@asu.edu} \\
   \And
 Sarath Sreedharan \\
  Department of CS, CSU \\
\texttt{sarath.sreedharan@colostate.edu} \\
\And
  Erin Chiou \\
  HSE, ASU \\
  \texttt{erin.chiou@asu.edu} \\
\And
  Subbarao Kambhampati \\
  SCAI, ASU \\
  \texttt{rao@asu.edu} }

\begin{document}
\maketitle

\begin{abstract}
Handling trust is a core requirement of effective interaction between people and AI agents. Thus, any decision-making framework designed to work with people must be able to estimate human trust. In this paper, we propose a mental model-based framework of trust that captures multidimensional aspects of trust and can be used to infer human trust. This framework can also be used as a foundation for any trust-aware decision-making framework. Through our mental model-based framework of trust, we propose a formal framework that captures intuitions about trust present in extant literature. We show how our framework captures various dimensions of trust perception and then use the framework to define trust evolution, human reliance, and decision-making. Additionally, we also propose a formalization of the appropriate level of trust in the agent.\\
  Using human subject studies, we evaluate if (1) changes in trust can be achieved by adjusting human beliefs about the agent according to predictions made by our mental model framework and (2) adjustments in human belief lead to corresponding modifications in trust perceptions; performance, process, and purpose, by controlling different aspects of the model.
\end{abstract}


\section{Introduction}
As AI-powered systems are becoming more pervasive in our day-to-day lives, the need to develop effective and intuitive interaction mechanisms between the human and the AI agent has become a pressing challenge.
A core mechanism that will determine the effectiveness of such interactions-- and in the long run the human's readiness to work with the system-- is the level of trust they place on the system.
Thus, understanding, estimating, and engendering an {\em appropriate level of trust} in the human end-users are challenging open problems that need to be addressed for meaningful deployment of AI systems in critical scenarios.
Here the qualification of engendering `{\em appropriate level}' is quite critical, as over trust or misplaced trust in the system could lead to critical problems such as automation bias and complacency \cite{parasuraman2010complacency}, with possibly detrimental performance outcomes. On the other hand, lack of trust can result in ignoring the system's capability and as a result lower team performance \cite{lee2004trust,chen2018planning,zahedi2021trust}.
Designing such a trust-aware AI system or framework would benefit from a generalizable and consistent formalization of trust that can be used in different decision-making frameworks. 

Prior research on trust in AI agents examined trust from human factors and computational trust frameworks. However, a deep dive into various dimensions and characteristics of trust, and the way they are treated in existing empirical studies including computational approaches reveals a significant gap in the literature. This gap lies in the inherent difference in the human factors view of trust, which views trust as a rich multi-dimensional phenomenon, versus the current computational approaches that usually conflate trust with reliability (only one of many dimensions discussed in the literature). This latter strategy is usually too coarse to capture all aspects of trust that could come into play within human-AI interactions.

For example, predicting trust in an AI solely based on its reliable performance neglects the complexity of human trust. Consider the AI achieving a goal through an unconventional and inexplicable process or pursuing goals not aligned with the user's preferences. Existing computational approaches may miss these complexities, highlighting the limitations of relying solely on performance metrics to understand trust in human-AI interactions.

Lee and See \cite{lee2004trust} characterize trust as based on different types of information that can guide expectations about the trustee's ability to help achieve the trustor's goals. These types of information fall into three categories: purpose, process, and performance \cite{lee2004trust}. We are unaware of any existing trust-aware decision-making methods that explicitly and completely account for these three types of trust information.

This paper provides a more comprehensive and consistent mental model-based framework of trust that can be used as a predictive method to infer trust, while also capturing the often-overlooked multidimensional aspects of trust, providing a foundation for the development of various trust-aware decision-making frameworks.

\begin{figure}
    \centering
    \includegraphics[width=.5\textwidth]{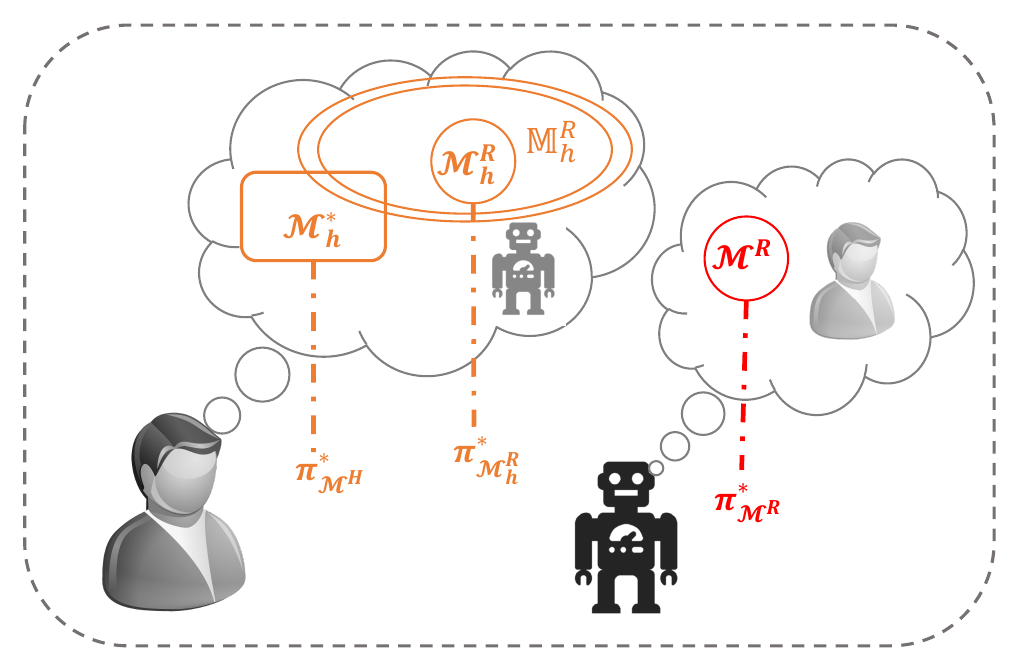}
    \caption{A schematic representation of the mental model based framework of trust. $\mathcal{M}^R$ is the task model that the AI agent ascribes to itself; $\mathbb{M}^R_h$ is set of models the human may ascribe to the agent; $\mathcal{M}^*_h$ is the human's task model that captures their expectations about the task.}
    \label{fig:base3model}
\end{figure}

We formalize a {\em mental model based framework of trust (MMbFT)} that frames human trust of an AI system in terms of their mental models and beliefs about a system. In this framework (see Figure~\ref{fig:base3model}), we will contextualize a person's trust and their consequent choice on whether or not to rely on the system in terms of their mental model of the agent (captured by a set of models $\mathbb{M}^R_h$) and their pre-existing expectations about the optimal way of solving the task (captured using the model $\mathcal{M}^*_h$). The agent on the other hand can either use behavior generated through its model (i.e. $\mathcal{M}^R$), or explanations about its model, to influence both a person's expectations about the agent and their expectations about the task. We use this framework to formalize the appropriate level of trust, model trust evolution and capture multi-faceted dimensions often overlooked in computational trust studies.

Using human subject studies we evaluate our mental model based framework of trust. We confirm that modifying individuals' beliefs about the agent, as per our mental model framework, results in changes in trust (as measured by a questionnaire).
Additionally, we study how different dimensions of trust (i.e., captured by performance, process, and purpose information and corresponding expectations) are affected by changes in human beliefs, and how well it aligns with the predictions made by our model.

\section{Related Work}
We examine trust from social psychological and human factors perspectives and computational trust models in human AI interaction, emphasizing the critical requirement for a generalized framework that bridges the gap between these treatments.

\textbf{Social Psychological and Human Factors Perspectives of Trust:}
Trust, a multidimensional and extensively explored concept, spans various research fields, including psychology, sociology, and human factors \cite{hoff2015trust}. In human factors, trust has been extensively studied, particularly in the context of guiding interactions with various technologies. Trust has been defined in diverse ways, ranging from an attitude or expectation to an intention or willingness to act \cite{rotter1967new,rempel1985trust,barber1983logic,johns1996concept,moorman1993factors,mayer1995integrative}. The most widely cited definition emphasizes vulnerability, with trust involving willingly taking risks by delegating responsibility to another party \cite{mayer1995integrative, rousseau1998not}. Moreover, trust can be seen as an outcome of behavior or a state of vulnerability or risk \cite{deutsch1960effect, meyer2001effects}. Among these definitions, Lee and See's definition stands out, defining trust as "the attitude that an agent will help achieve an individual's goals in a situation characterized by uncertainty and vulnerability" \cite{lee2004trust}. Therefore, Lee and See considers the basis of trust as information-based, guiding expectations regarding how well the trustee can achieve the goals. This information supports three levels of attributional abstraction: purpose, process, and performance information, which describe the agent's goal-oriented characteristics \cite{lee2004trust}.

In quantifying and measuring trust, various instruments such as the Muir questionnaire assess trust in terms of reliability, predictability, faith, and overall trust \cite{muir1994trust}. Extensions by \cite{sugcummings2008selecting}, \cite{2ullman2018does}, and \cite{82jian2000foundations} include factors like perceived reliability, technical competence, understandability, faith, and personal attachment. Some researchers also explore self-reported trust, considering predictors such as propensity to trust machines and attitudes toward automation \cite{81merritt2013trust}. A wide range of trust assessment methods have been used including myriad bespoke questionnaires \cite{Alsaid_Li_Chiou_Lee_2023} and indirect measures using physiological data \cite{Kohn_deVisser_Wiese_Lee_Shaw_2021}.

\textbf{Computational Approaches to Trust:} Trust in human robot interaction (HRI) encompasses trust inference, trust utilization, trust calibration and repair. Trust inference leverages observed human behavior to predict trust levels \cite{9-desai2012modeling}. One notable model, the Online Probabilistic Trust Inference Model (OPTIMo), captures trust as a latent variable within a dynamic Bayesian network \cite{xu2015optimo}. 

Bayesian inference and non-parametric methods, such as Gaussian processes and Recurrent Neural Networks (RNN), have further advanced trust inference \cite{soh2020multi}. 
Trust utilization involves integrating estimated trust into mechanisms that modify robot behavior to enhance collaboration efficiency \cite{nikolaidis2017human,zahedi2023trust,wang2018trust,chen2018planning}. Trust calibration and repair address adjusting and restoring trust levels, including methods like apologies and transparency enhancements \cite{baker2018toward,tolmeijer2020taxonomy,de2018automation,sebo2019don,44-wang2016trust}.

\textbf{Bridging the Gap} Despite advances in both human factors and computational trust models in HRI, a noteworthy gap exists between these perspectives. Human factors research delves into multifaceted trust dimensions, focusing on human cognition and behavior. In contrast, computational models primarily address trust as reliability, often through simplified task contexts that miss key trust dimensions. While Bayesian trust models attempt to derive multi-aspects of trust from evidence \cite{thirunarayan2014comparative,ries2009extending,azevedo2021unified,rodriguez2023review,guo2021modeling}, their application-specific nature, reliance on independence assumptions, and computational challenges limit their adaptability and practical utility. Furthermore, the lack of generalizability in these models may in part be due to oversight of certain important dimensions of trust. This oversight then constrains their ability to capture the underlying complexity of trust, limiting its subsequent predictability in various settings. For example, researchers have proposed Bayesian trust models that account for both internal and external attributes of trust, such as reputation and behavior \cite{teacy2012efficient,wang2003bayesian,castelfranchi2003trust} or different social dimensions of trust \cite{melaye2005bayesian}. However, these efforts highlight the ongoing challenge of achieving a fully comprehensive understanding within the confines of the Bayesian framework. 

In contrast, our mental model-based framework serves as a versatile foundation, allowing for the incorporation of multiple dimensions and information types of trust, including but not limited to performance, process, and purpose. Unlike Bayesian models, our framework provides a customizable architecture for adapting and extending to specific trust dimensions relevant to the application. For instance, in HRI scenarios requiring control over one or many specific perceptions of trust—such as the ones related to task performance, process efficiency, and alignment with user goals. Moreover, it acts not only as a trust inference tool but also as a predictive model, anticipating changes in various perceptions of trust. This unique adaptability and predictive capability make our framework a powerful and flexible tool for trust modeling in diverse contexts, serving as a solid base for the development of comprehensive decision-making frameworks.

\section{Interaction Setting}
Our test scenario consists of two agents types, an AI agent (denoted by $R$) and a human agent trying to derive useful decisions by observing the agent (denoted by $H$). We will model each agent that is part of the scenario as a model-based agent: the agent uses a model to not only derive its own decisions but to form expectations about what other agents might or should do. Note that we use the term ``model'' loosely to mean any formal model that encodes, among other things, an agent's beliefs about task objectives, state of the world and how the world may evolve on its own or in response to an agent action. There exist a wide variety of options for what the models might be (for example they may be MDPs \cite{puterman2014markov}, POMDP or symbolic models like PDDL \cite{geffner2013concise}). We only require that the models used by an agent are in a form that can be used to derive the required decision or expected decisions.

As apparent from Figure \ref{fig:base3model}, our framework of trust consists of the interaction between three sets of models. We will provide a detailed description of each type of model and our theory of trust.\\

\noindent\textbf{Model of the agent $\mathcal{M}^R$:} This is the model the AI or robot agent ascribes to itself. This determines what actions the agent "believes" it could perform and the objectives and preferences it is trying to satisfy.\\

\noindent\textbf{Set of the human's models of the agent $\mathbb{M}^R_h$:} This set consists of the models a person may ascribe to the robot agent, such that each model $\mathcal{M}^R_h \in \mathbb{M}^R_h$ is, as far as the person is concerned, could be the actual model being used by the agent to derive its decisions. Each model is also associated with some likelihood ($P_{\mathbb{M}}$) that corresponds to the person's degree of certainty that a specific model corresponds to the true model used by the agent.\\

\noindent\textbf{Human's task model $\mathcal{M}^*_{h}$:} This model is meant to capture the expectations the person may have about the task that is independent of what they believe the agent is capable of performing. In our framework of trust, this model mainly serves to capture the person's expectations of the idealized way to complete the task. Such expectation may be formed from beliefs about their own ability to complete the task or may even come from other sources (either from observing more experienced users, or the expectations may be carried over from institutional expectations). One important point to note here is that this is the person's belief about the ideal way of completing the task and need not reflect the true optimal ways of completing the task for the robot agent. Additionally, to allow Bayesian reasoning, we will assume that at least some of the models in the set $\mathbb{M}^R_h$ can generate solutions comparable to those that are generated by $\mathcal{M}^*_{h}$, however, the prior likelihood on those models may be quite low.

As discussed, the models referred to in our framework can, in theory, be any decision-making model. However, to provide a grounded technical exposition, we will assume all models discussed are captured using STRIPS representation \cite{geffner2013concise}.
In such cases, a model may be represented as a tuple of the form $\mathcal{M} = \langle \mathcal{D}, \mathcal{I}, \mathcal{G}\rangle$, consisting of domain $\mathcal{D} = \langle \mathcal{F}, \mathcal{A} \rangle$, where $F$ being a finite set of fluents that define the state of the world $s\subseteq \mathcal{F}$, $\mathcal{A}$ the finite set of actions, $\mathcal{I}$ the initial state, $\mathcal{G}$ the goal, where $\mathcal{I}, \mathcal{G}\subseteq \mathcal{F}$. 
A plan for such models is a sequence of actions $\pi = \langle a_1,...,a_k \rangle$, which when executed in the initial state leads to the goal $\rho_{\mathcal{M}}(\mathcal{I}, \pi)$ $\models$ $\mathcal{G}$. The cost of a plan $\pi$ is given by $C(\pi,\mathcal{M}) = \sum _{a\in \pi} c_a$ if $\rho_{\mathcal{M}}(\mathcal{I}, \pi)$ $\models$ $\mathcal{G}$, otherwise the cost is $\infty$. The likelihood of a plan given a model will take the form $P_l : \mathcal{M} \times \pi \rightarrow [0, 1]$. While we will try to be agnostic to
likelihood functions, a fairly common approach \cite{fisac2018probabilistically,baker2009action} is a noisy rational model based on the Boltzmann distribution: $P_l(\mathcal{M},\pi) \propto e^{-\beta \times C(\pi)}$ \cite{sreedharan2020bayesian}. The cheapest plan with highest probability is called optimal plan $\pi^* = arg \min_{\pi}C(\pi,\mathcal{M})$ $\forall \pi$ such that $\rho_{\mathcal{M}}(\mathcal{I}, \pi)$ $\models$ $\mathcal{G}$.

\section{Human's Trust and Their Reliance}
With the basic interaction setting in place, we are ready to provide a preliminary definition of trust and a way to model a person's choice to rely on a specific AI agent. 

\paragraph{Trust} 
We will follow the recent definitions of trust \cite{jacovi2021formalizing} that tend to focus on the fact that trust is a contextual measure. This argues that trust as a measure is best understood in terms of one's expectation on an agent to satisfy some specific contract as opposed to thinking of it as a general measure associated with the agent. In our model, the contract corresponds to the agent's behavior meeting some specific performance guarantee (generally related to the quality of the solution). This contract, denoted as $\mathcal{C}$, will be formed by the human based on their model $\mathcal{M}^*_h$. 

\begin{figure}
    \centering
    \includegraphics[width=.25\textwidth]{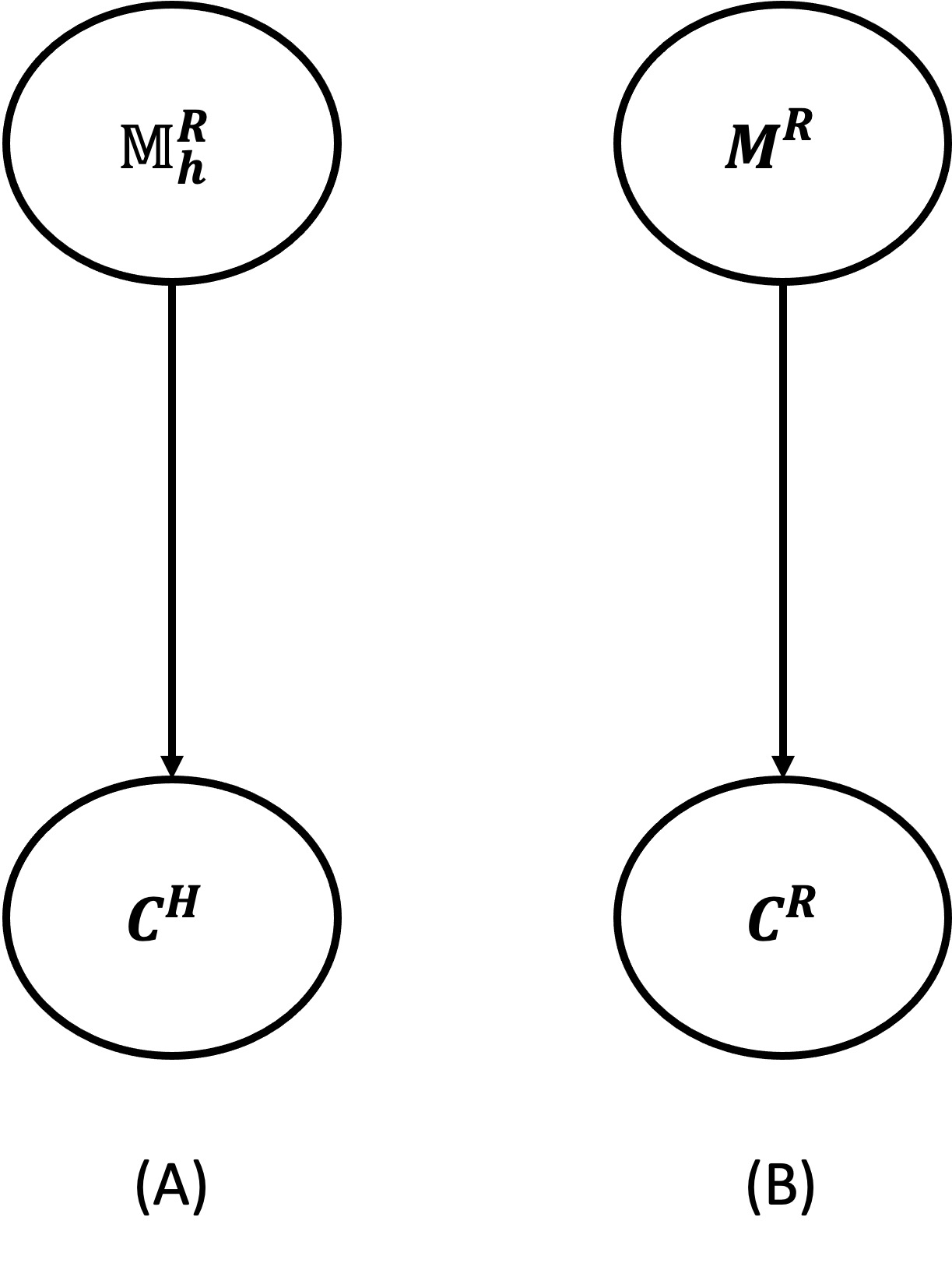}
    \caption{A graphical model representing the probabilistic reasoning that is performed in this setting: Subfigure (A) captures the reasoning performed at the human's end with $\mathcal{C}^H$ being the random variable corresponding to the human's belief that a contract $\mathcal{C}$ will be satisfied. Similarly subfigure (B) represents the reasoning performed at the human's end, where $\mathcal{C}^R$ captures whether the robot achieves the contract $C$}
    \label{fig:prob_model}
\end{figure}

Now the {\em trust measure} ($\mathcal{T}(\mathcal{C})$), i.e., the numeric quantity reflecting the degree of trust the human places on the agent in the context, will be defined to be directly proportional to a monotonically increasing function over the likelihood the human places on the agent to satisfy the contract, i.e.,
\[\mathcal{T}(\mathcal{C}) \propto \mathcal{F}(P(\mathcal{C}^H))\]
Where the likelihood the person would believe the agent can satisfy the contract $P(\mathcal{C}^H)$ ($\mathcal{C}^H$ is the random variable corresponding to the human belief that the contract will be satisfied) is, in itself, controlled by the human's belief about the agent's model. Specifically, we will assume that the human reasoning can be captured using the probabilistic graphical model presented in Figure \ref{fig:prob_model}(A) and the likelihood is given as

\[P(\mathcal{C}^H) = \sum_{\mathcal{M}\in \mathbb{M}^R_h} P(\mathcal{C}^H|\mathcal{M}) \times P_{\mathbb{M}}(\mathcal{M}) \]
Where $P(\mathcal{C}^H|\mathcal{M})$ is the likelihood human associates with a model $\mathcal{M}$ coming up with a solution that can satisfy a given contract $\mathcal{C}$.\\

Regarding the likelihood that the person associates with a model coming up with a solution that can satisfy a given contract, we can assume it is a monotonic function over the cost of the optimal plan in that model. A common function one could adopt is a Boltzmann distribution over the cost of the optimal plan in that model.

\paragraph{Three Information Types of Trust Framework}
In human factors research, it is suggested that appropriate trust hinges on the understanding of an agent's performance, process, and purpose. Thus, it is paramount for a computational trust model to encompass these three information types in the task environment. Our mental model-based framework is designed to consider these information types that affect trust. Here, we dive deeper into the alignment of these trust information types with our mental model-based framework. Initially, we will define each information type and then illustrate how they are captured in our model.

In our framework, it is important to recognize that the formation of trust in humans is tightly bound with their mental models. So, the trust a human places in a robot agent is fundamentally linked to the interplay between the human's mental models ($\mathcal{M}^*_h$ and $\mathbb{M}^R_h$). To logically reason about trust, we must therefore focus our attention on the section of the framework capturing all models related to the human.

To delve deeper into the mechanics of our model, consider one of the human's model of the robot agent that belongs to the set of human models of the robot, $\mathcal{M}^R_h \in \mathbb{M}^R_h$. This model is represented as $\langle \mathcal{D}^R_h, \mathcal{I}^R_h, \mathcal{G}^R_h\rangle$. Here, $\mathcal{D}^R_h$ symbolizes the domain of the model, $\mathcal{I}^R_h$ refers to the initial conditions, and $\mathcal{G}^R_h$ encapsulates the goal state. This goal state, $\mathcal{G}^R_h$, comprises both the internal AI agent's goal and the goal as articulated by the human agent.

With this understanding, we can now map these components of our model to the three information types of trust:
\begin{enumerate}
    \item \textbf{Performance:} Performance information depicts the AI agent's functions and speaks to the competency or proficiency of the AI agent as indicated by its success in fulfilling human objectives \cite{lee2004trust}. This information corresponds to the domain of the model, $\mathcal{D}^R_h$.

    \item \textbf{Purpose:} Purpose information elucidates why the AI agent was conceived and assesses whether the agent has a positive orientation towards the human agent \cite{lee2004trust}. This information is associated with $\mathcal{G}^R_h$ within our model, which consists of both the AI agent's internal goal and the human agent's specified goal.

    \item \textbf{Process:} Process information explores the operational procedures of the AI agent and evaluates the quality of actions based on human expectations \cite{lee2004trust}. It correlates with the extent to which a plan derived from the model matches the person's desired plan. To quantify this alignment, we can use cost as a metric. Thus, \textit{process} can be seen as the probability that a model's optimal solution aligns with the human's optimal solution, $\pi^*_{\mathcal{M}^*_h}$. This probability is mathematically defined via the Boltzmann distribution of the cost difference between the two plans, expressed as $P_e(\mathcal{M}^R_h, \pi^*) \propto e^{-\beta(C(\pi^{\mathcal{M}^*_h})-C(\pi^*)}$.
\end{enumerate}

\paragraph{Modeling Trust Evolution}  
As a person's trust is directly tied to their uncertainty about the agent model, their trust evolves as their belief over $\mathbb{M}^R_h$ changes. As discussed earlier, we will be modeling people as Bayesian reasoners in this context, and as they observe the AI agent's behavior, the person would be expected to update their posterior over the models the agent may hold. In particular, this belief evolution is expected to result in an increase in trust,
if the updated model distribution causes
the likelihood $P(\mathcal{C}^H)$, and by extension the trust measure $\mathcal{T}(\mathcal{C})$ to increase. Because the contract itself is drawn from the model $\mathcal{M}^*_h$, such increases in trust are usually achieved by placing more probability in models that are closer match with $\mathcal{M}^*_h$. Another way $\mathbb{M}^R_h$ could be updated could be through {\em explanations}.\\
Therefore, our mental model framework can predict both the dynamics of trust changes and shifts in specific perceptions of trust. In the upcoming evaluation section, we'll assess these predictions, providing a clear understanding of the effectiveness of our model in inferring trust evolution.
 
\paragraph{Formalizing Appropriate Levels of Trust} When one speaks of developing a framework to formalize trust and methods to engender trust, the common concerns raised are the ones related to {\em automation bias} and {\em automation complacency} \cite{parasuraman2010complacency}. 

Each of these scenarios is characterized by the human agent placing an unwarranted amount of trust in the agent's capability, with possibly disastrous consequences. With our more formal grounding and definition of trust, we are now capable of formalizing what it means to engender an appropriate level of trust to avoid such issues. In particular, we can assert that the level of trust a person has in an agent with respect to a contract is appropriate if the likelihood the person associates with the system satisfying the contract is equal to the likelihood of the agent satisfying that contract, i.e.,
{\em trust level is appropriate}, if

\[
P(\mathcal{C}^H) = 
\sum_{\mathcal{M}\in \mathbb{M}^R_h} P(\mathcal{C}^H|\mathcal{M}) \times P_{\mathbb{M}}(\mathcal{M}) =
P(\mathcal{C}^R|\mathcal{M}^R).
\]
Where $\mathcal{C}^R$ is the random variable associated with the robot actually achieving the contract (see Figure \ref{fig:prob_model} (B)).

\paragraph{Modeling Human's Reliance} One of the important behaviors that we are interested in capturing is whether the human agent is ready to accept the current decision given their trust in the agent. By grounding trust in terms of likelihood of goal achievement, we are now able to leverage decision-theory to model the expected values of the person choosing to accept ($Acc$) or not accept ($\neg Acc$) a given decision. For the choice $Acc$, there are two possibilities, the decision in fact satisfies the contract and thus receives some positive utility for being successful  ($U_{+\mathcal{C}}$) or it may fail in which case it gets a negative utility as a penalty ($U_{-\mathcal{C}}$). The expected value of relying on the AI agent's decision is thus given as 
\[V^{\mathcal{C}}(Acc) = P(\mathcal{C}^H)*U_{+\mathcal{C}} + (1- P(\mathcal{C}^H))*U_{-\mathcal{C}}\]
While the value of choosing to not accept the AI agent is basically given as 
\[V^{\mathcal{C}}(\neg Acc) =-1 * C_{\neg Acc}  \]
Where $C_{\neg Acc}$ is the penalty associated with the person choosing to turn down a decision from the agent.  
The person would accept a given decision if $V^{\mathcal{C}}(Acc) > V^{\mathcal{C}}(\neg Acc)$. 

\section{Evaluation of Trust Inference Through Human Subject Experiments}

In this section, we assess our proposed theory by comparing it to a trust questionnaire developed by Chancey et al. \cite{chancey2017trust}. The central focus of our framework that we aim to examine pertains to the potential for inducing changes in trust (as measured through the questionnaire) by altering the human's belief about the agent, as predicted by our mental model theory.

In particular, we divide our participants into two groups: a positive update group, in which participants' initial belief in the agent's capacity to fulfill a specific contract is increased during the study, and a negative update group, in which participants' beliefs are decreased. In the course of evaluating our theory, we first investigate whether these two distinct belief updates result in different levels of trust. Once this distinction is established, we explore whether the positive update group experiences an increase in trust, and whether the negative update group experiences a decrease in trust.

Moreover, we examine whether modifying a specific component of the model leads to a corresponding alteration in the related information aspect of trust; performance, process, and purpose. We consider three scenarios in which we update specific parts of the model associated with performance, process, and purpose. Subsequently, we assess whether a positive or negative update in a specific component of the model influences an increase or decrease in the associated perception of trust, respectively.

Our expectation is that when the update corresponds to a specific aspect of trust information, it will primarily impact the perception of trust associated with that particular aspect, resulting in a stronger correlation than with other information types of trust.

The specific hypotheses we aim to test are as follows:
\begin{description}
    \item [H1-] The change in trust induced by an increase in likelihood of satisfying the contract is different from the one induced by a reduction in the likelihood of satisfying the contract.
    \item[H2-] The positive update group will exhibit an increase in trust, as measured by a trust questionnaire, compared to their initial trust levels.
    \item[H3-] The negative update group will demonstrate a decrease in trust, as measured by a trust questionnaire, in comparison to their initial trust levels.
    \item[H4-] Positive/negative update of the model corresponding to the performance will increase/decrease performance perception of trust more than other perceptions.
    \item[H5-] Positive/negative update of the model corresponding to the process will increase/decrease process perception of trust more than other perceptions.
    \item[H6-] Positive/negative update of the model corresponding to the purpose will increase/decrease purpose perception of trust more than other perceptions.
\end{description}
\subsection{Experiment Setup} 
We conducted a between-subjects study design to assess the impact of either a positive or negative update, along with the specific type of update related to performance, process, or purpose, on participant trust. As a result, each participant encountered one of the following combinations: positive or negative update, and one of the three scenarios: performance, process, or purpose.
\subsubsection{Events} 
Each participant experiences two key events. In the initial event, they are presented with a task description and information about potential robots they might encounter. These robots vary in their capabilities, and participants are provided with images displaying the robot within its task environment. Following this, participants' trust is assessed using the trust questionnaire.

In the subsequent event, participants view a video showcasing a robot engaged in performing the task. This video serves to update their mental model of the robot's capabilities when executing the task. Following this update, their trust is reassessed. The first event aims to establish  participants' initial mental models of the robot ($\mathbb{M}^R_h$), while the second event serves to update those mental models ($\mathbb{M}^R_h$).
\begin{figure*}[t]
\begin{subfigure}[\label{fig:performance}]
    \centering
    \includegraphics[width=0.3\textwidth]{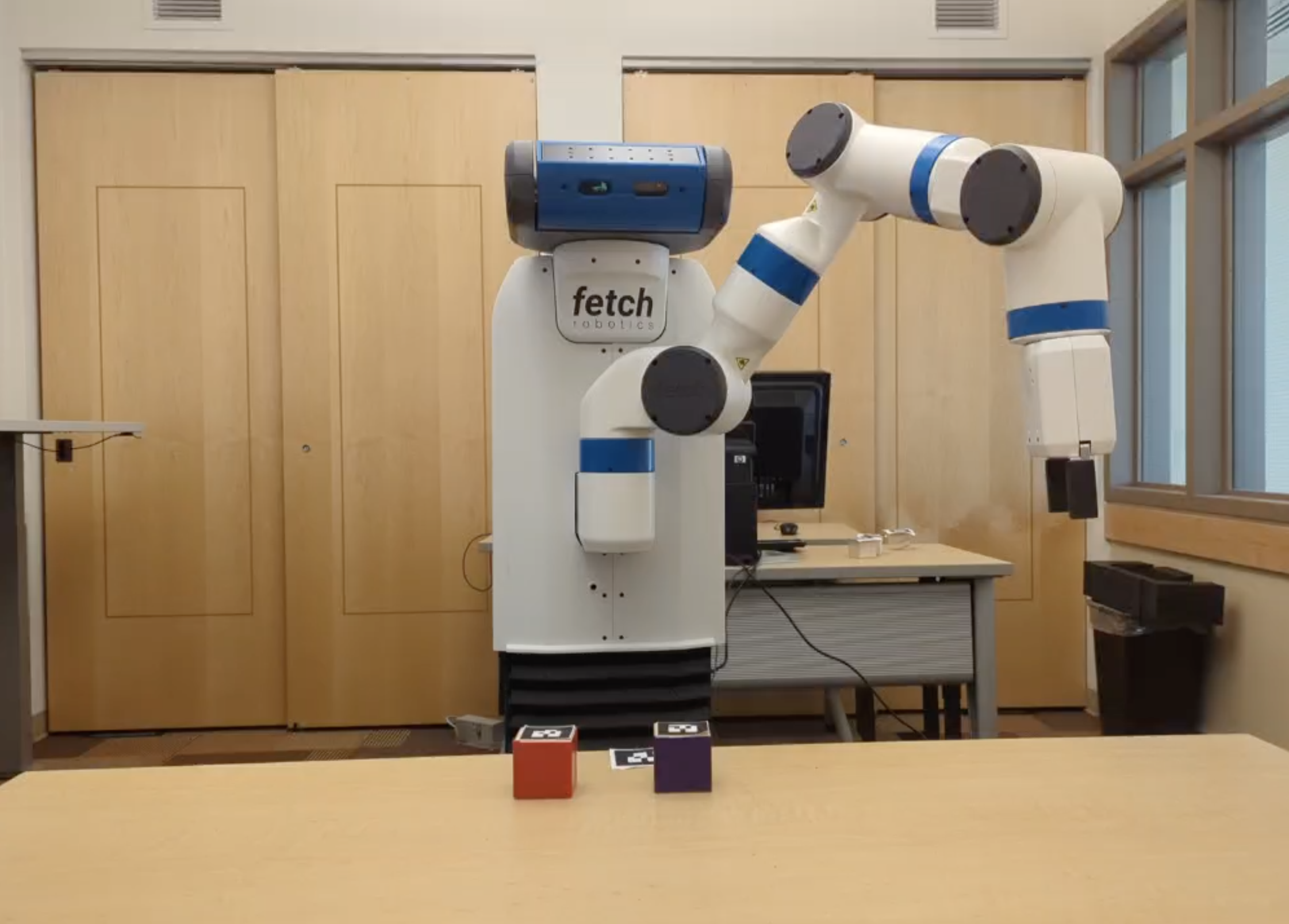}
    \hfil
    \end{subfigure}
    \begin{subfigure}[\label{fig:process}]
    \centering
    \includegraphics[width=0.3\textwidth]{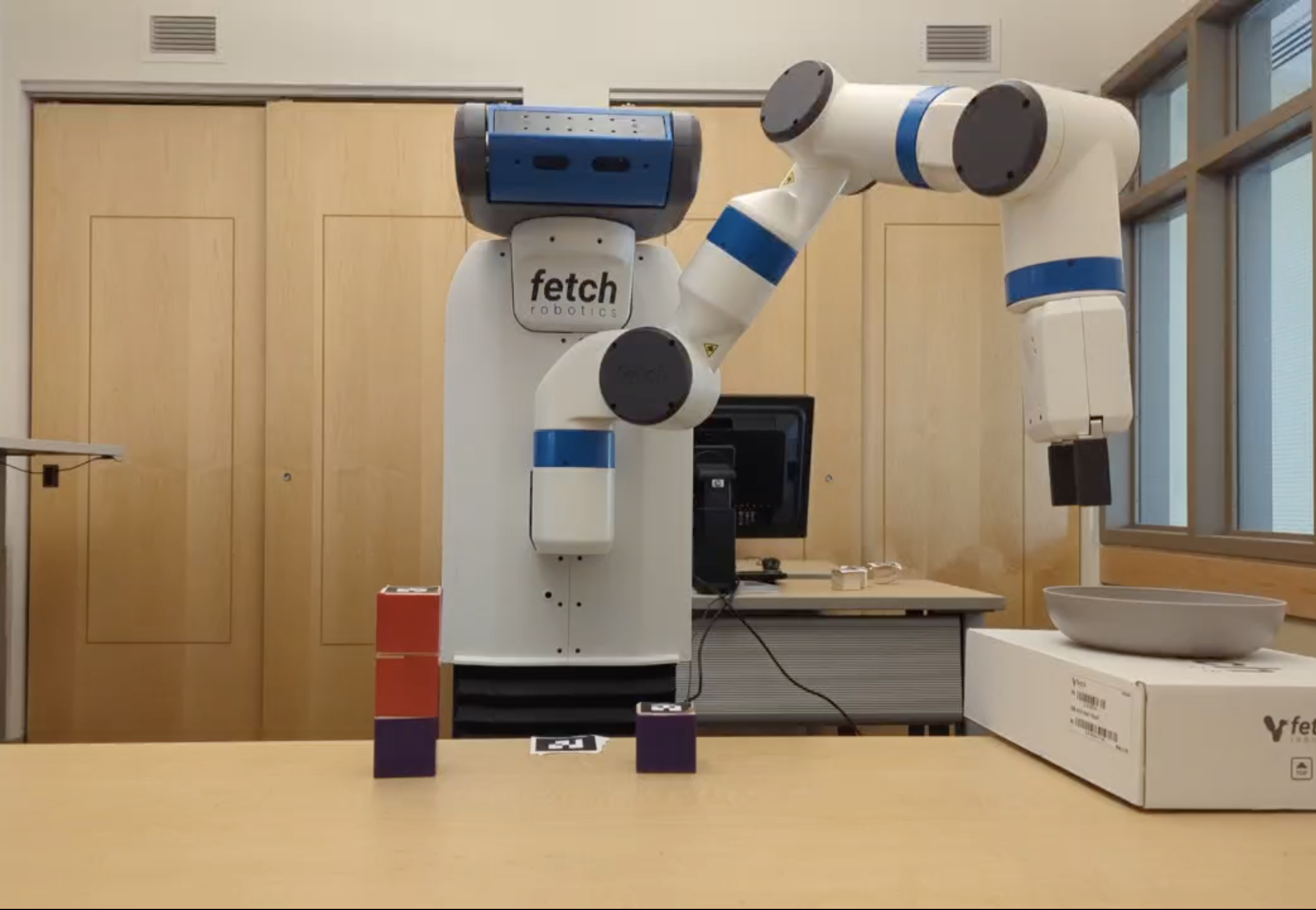}
    \hfil
    \end{subfigure}
    \begin{subfigure}[\label{fig:purpose}]
    \centering
    \includegraphics[width=0.3\textwidth]{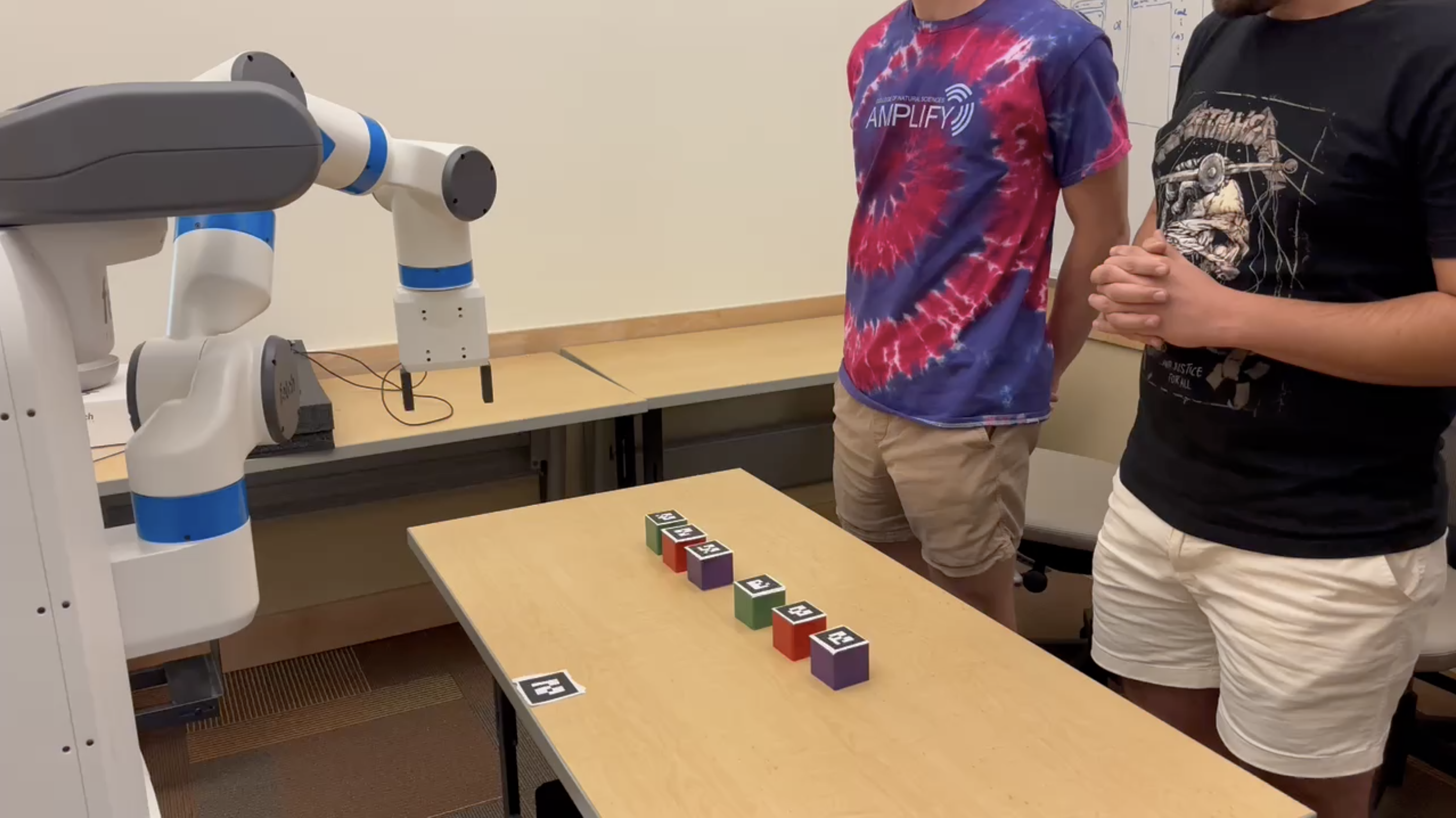}
    \hfil
    \end{subfigure}
    \caption[The Robot in Its Task Environment in the Three Scenarios (a) Performance Scenario (S1), (b) Process Scenario (S2) And (c) Purpose Scenario (S3).]{The robot in its task environment in the three scenarios (a) Performance Scenario (S1), (b) Process Scenario (S2) and (c) Purpose Scenario (S3).}
    \label{fig:scenario}  
\end{figure*}
\subsubsection{Scenarios} We have three scenarios in our studies. Each of the scenarios has both positive and negative updates that make six cases for our between-subjects study. The scenarios are:

\textbf{(1) Performance Scenario (S1):} In this scenario, we manipulate the performance information of the model to update the human model of the robot. The robot's task is to place the purple block on the red block, the robot in its task environment is illustrated in Figure \ref{fig:performance}. Initially, participants receive information about the task setting and details about the robots, which form their initial mental model of the robot. In this scenario we inform participants that (1) the purple block is heavy, and (2) uncertainty regarding the type of robot they will encounter. Equally likely, participants may be exposed to one of the following robots: \\(a) A robot that is capable of picking up heavy blocks.\\
(b) A robot that is not able to pick up heavy blocks.

Following the initial event, participants receive either a positive or negative update regarding the robot's performance through a video demonstration of the robot performing the task. In the positive update, the robot successfully lifts the purple block and places it on the red block. In contrast, the negative update features a robot attempting to lift the purple block but failing to do so, ultimately ceasing the task after an unsuccessful attempt.

\textbf{(2) Process Scenario (S2):} Similarly, this scenario involves a condition where we manipulate process information within the model to update the human model of the robot. In this scenario, the robot's task is to place a purple block in a bowl. The robot in its task environment is depicted in Figure \ref{fig:process}. Participants receive information about the task setting and details about the robots, which form the basis of their initial mental model of the robot. In this scenario, we inform participants that (1) there are two purple blocks, and two red blocks, as visually represented in the image. (2) the robot should always try to complete its task in the easiest way.

Following this initial event, participants are presented with a video update that showcases the robot's approach to the task. In the positive update, the robot efficiently selects the accessible purple block and places it into the bowl. Conversely, the negative update portrays a different approach: the robot initially chooses to pick up the red blocks and place them on the table. It then proceeds to pick up the purple block that was beneath the red blocks before ultimately placing it in the bowl.

\textbf{(3) Purpose Scenario (S3):} In this scenario, we focus on manipulating the purpose-related information within the model to update the human model of the robot. The robot's task involves a unique setting. Participants are informed that there are two individuals engaged in a block stacking competition, with each person having three blocks. The objective is for one person to stack their blocks before the other. Importantly, the participants are made aware that the players can only start stacking their blocks once the robot completes its operations. In other words, if the robot chooses to assist one of the players in stacking their blocks, that player is guaranteed to win the competition.
Participants are further informed that the person on the left, identified by wearing a black T-shirt, is their friend, and they wish for the robot to help their friend win in the competition. Consequently, their task for the robot is to support their friend by stacking the blocks for him. The robot, depicted in its task environment in Figure \ref{fig:purpose}, is accompanied by initial information provided to participants about the robots, shaping their initial mental model of the robot. In this scenario, participants are informed that it is equally likely that they may encounter one of the following types of robots:\\
1. Satisfying the user's goal is the robot's objective.\\
2. Satisfying the user's goal is not the robot's objective.

Following this initial event, participants receive either a positive or negative update related to the robot's purpose through a video demonstration of the robot's task. In the positive update, the robot assists their friend in stacking the blocks. In contrast, the negative update shows the robot helping the other person (their friend's rival). 

\subsubsection{Trust Questionnaire}
We employ a trust questionnaire developed by Chancey et al. \cite{chancey2017trust}. This questionnaire is explicitly designed and previously validated to assess three information types of trust: purpose, process, and performance. The original questionnaire was crafted within the context of an operator and a recommendation system, specifically, a tank-spotting aid. To adapt the questionnaire for use in our domain, we  carefully revised the questions to make sense within the context of our task environment, while capturing the core essence of the original questionnaire. The revised questionnaire used in this study is provided as supplementary material.

\subsubsection{Procedure}
Upon granting their consent, each participant is directed to the instruction page. Within these instructions, we introduce a gamified element to the study. Participants assume the role of technology purchasers tasked with making a critical recommendation to the CEO of their company regarding the purchase of a robot. The success of their company's future depends on their final decision. A mistake in their recommendation can reflect poorly for their standing as an employee within the company. We also provide a brief overview of the study's process.\\
Subsequently, participants are presented with a description of the task that the robot is expected to perform, accompanied by an image depicting the robot within its task environment. This image includes annotated objects for clarity. Furthermore, participants receive information about the robot and the setting, aligning with the scenarios outlined in the study. Following this phase, participants are asked to complete the fifteen-item trust questionnaire, which is accompanied by two attention check questions, all presented in random order.\\
Next, participants view a video showcasing the robot's execution of the task, after which they are prompted to respond to a similar set of questions. This phase requires them to consider the observations made during the video presentation.\\
Subsequently, participants are invited to provide their recommendation regarding whether their company should proceed with the purchase of the robot or not.\\
Lastly, participants are asked to complete a set of demographic questions to assess the population sample represented in this study.

\subsubsection{Participants}
A total of $123$ participants were recruited through the Prolific platform\footnote{\href{https://www.prolific.co/}{{https://www.prolific.co/}}}. These participants were randomly distributed among six cases, with $21$ participants in the Performance Scenario with Positive Update (S1PU), $21$ participants in the Performance Scenario with Negative Update (S1NU), $22$ participants in the Process Scenario with Positive Update (S2PU), $21$ participants in the Process Scenario with Negative Update (S2NU), $20$ participants in the Purpose Scenario with Positive Update (S3PU), and $18$ participants in the Purpose Scenario with Negative Update (S3NU). Each participant received $\$2$ as compensation for their participation.\\
The median time taken by participants to complete the study was $6$ minutes and $14$ seconds. In terms of gender distribution, $47.97\%$ identified as women, $49.59\%$ as men, $1.62\%$ as non-binary, and one person preferred not to specify their gender. Participants' ages were distributed as follows: $39\%$ were between the ages of $25$ and $34$, $17.89\%$ fell within the $35-44$ age group, $16.26\%$ were in the $18-24$ age range, $13\%$ were aged $45-54$, $10.57\%$ were in the $55-64$ category, and $3.25\%$ were $65$ or older.
Notably, the majority of participants ($60.97\%$) indicated some level of familiarity with AI and robotics.

\subsection{Results}
Across all six cases where we collected both initial and updated trust perceptions, we calculated participants' total trust at each step by computing the average across all fifteen questions. Additionally, we computed each perception of trust (performance, process, and purpose) separately by calculating the mean value for that specific perception.\\
Our data were grouped based on two criteria: (1) positive or negative update, which involves considering all data together for each update, and (2) data related to each scenario, considered separately for each update.\\
In addition to examining basic statistics related to the collected data, we also perform repeated measures t-tests and one-way ANOVA tests to test our hypotheses for each scenario as well as across all scenarios. We used a two-tailed t-test for the first hypothesis and a one-tailed t-test for the subsequent hypotheses. Any claims of statistical significance will be made against a significance level of $0.05$. The complete detail of the data from the statistical tests are provided in the supplementary material.

\begin{figure*}[t]
\begin{subfigure}[\label{fig:ttrust_per}]
    \centering
    \includegraphics[width=0.23\textwidth]{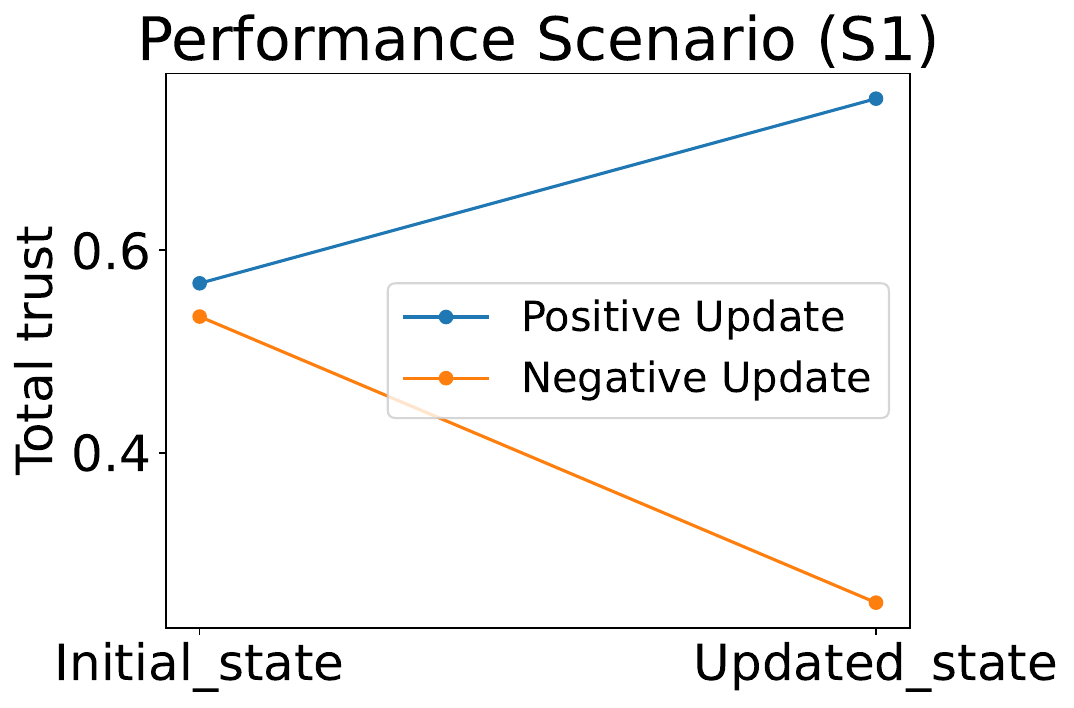}
    \hfil
    \end{subfigure}
    \begin{subfigure}[\label{fig:ttrust_pro}]
    \centering
    \includegraphics[width=0.23\textwidth]{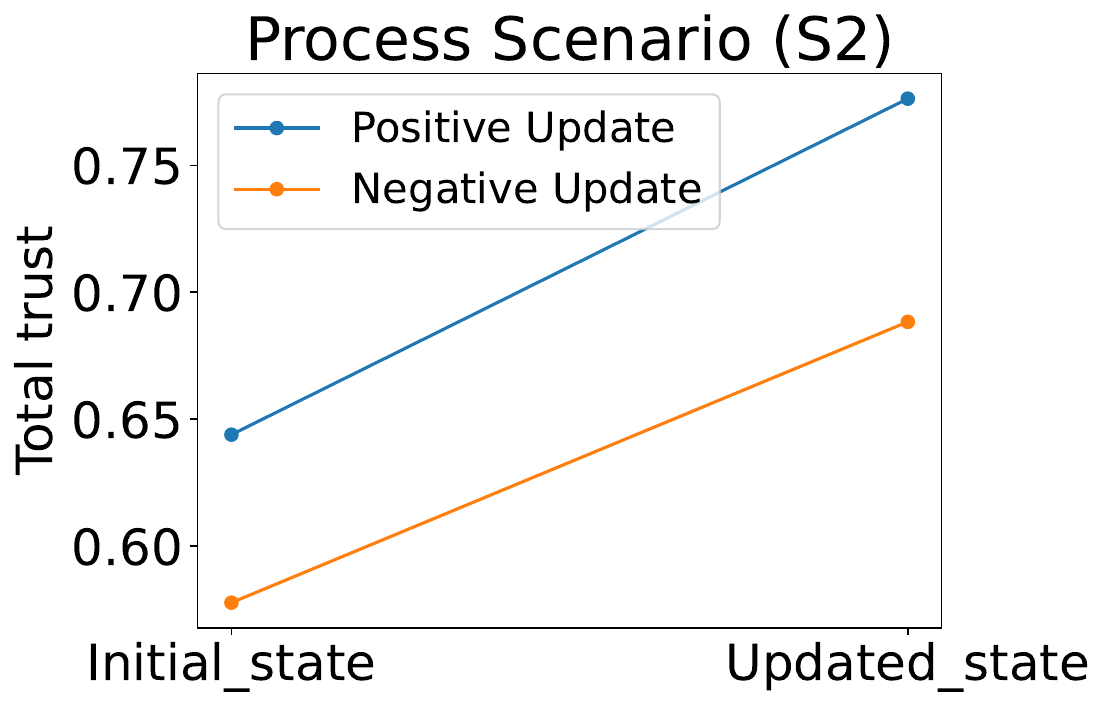}
    \hfil
    \end{subfigure}
    \begin{subfigure}[\label{fig:ttrust_pur}]
    \centering
    \includegraphics[width=0.23\textwidth]{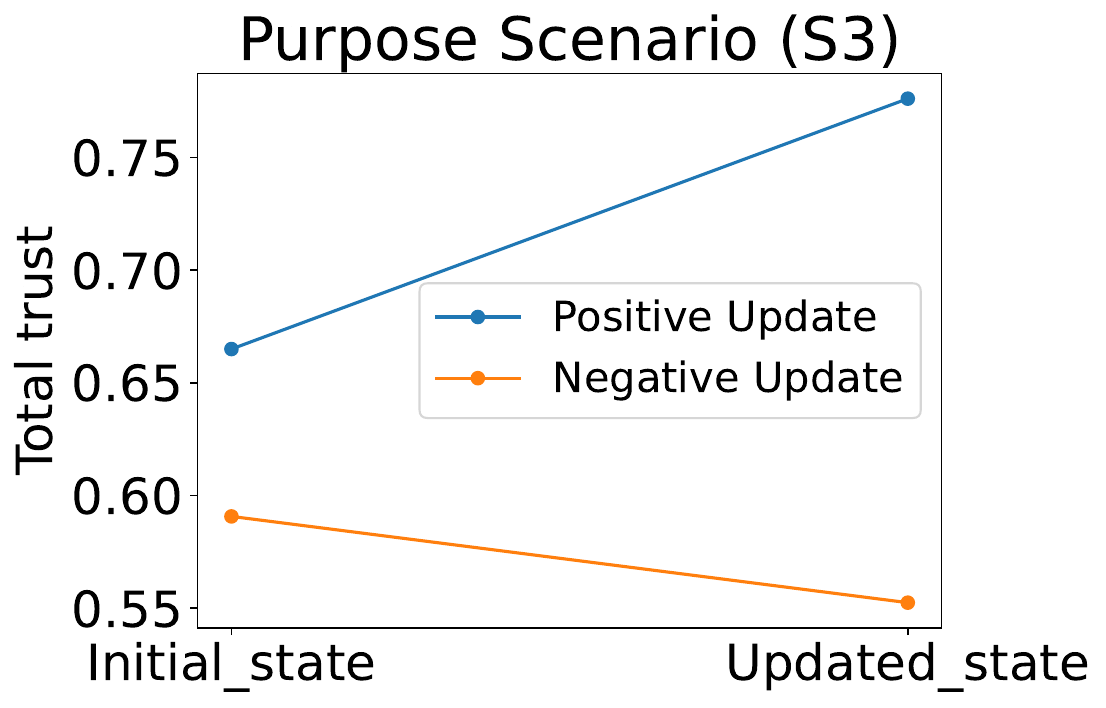}
    \hfil
    \end{subfigure}
    \begin{subfigure}[\label{fig:ttrust_all}]
    \centering
    \includegraphics[width=0.23\textwidth]{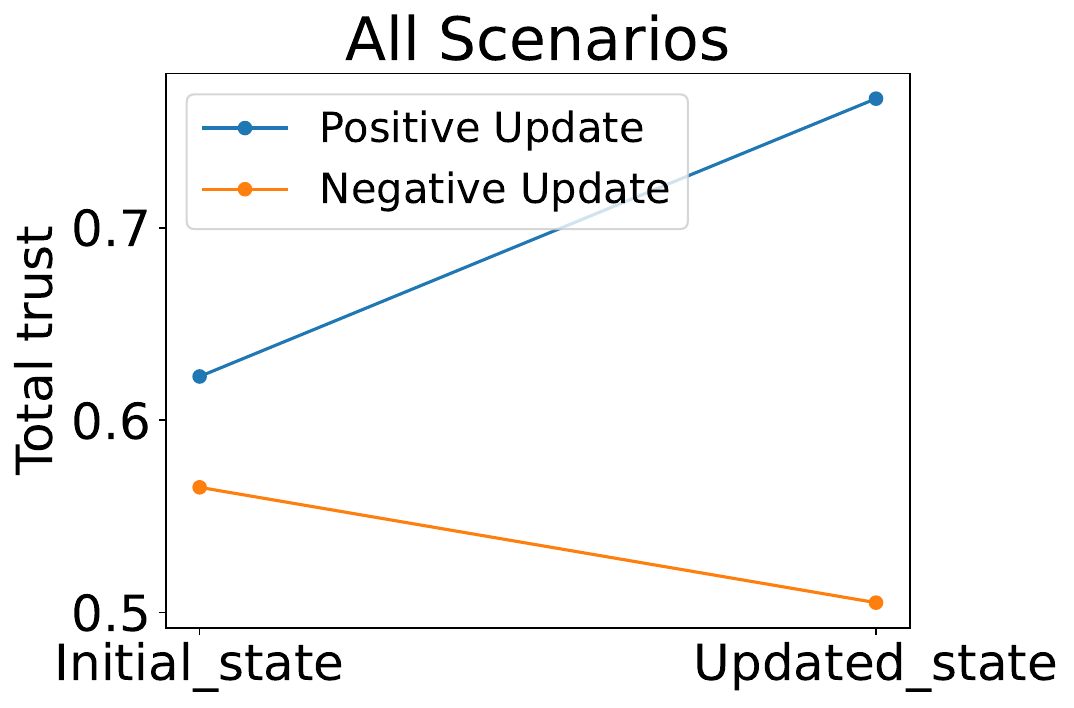}
    \hfil
    \end{subfigure}
    \caption[Total Trust Change From Initial State to Update State With Positive and Negative Updates. (a) Performance Scenario (S1), (b) Process Scenario (S2), (c) Purpose Scenario (S3) And (d) All Scenarios.]{Total trust change from initial state to update state with positive and negative updates. (a) Performance scenario (S1), (b) Process scenario (S2), (c) Purpose scenario (S3) and (d) All scenarios.}
    \label{fig:total_trust}  
\end{figure*}

\paragraph{H1- The change in trust value in positive vs. negative update}
In this analysis, we compare the updated trust values between the positive update and negative update groups. This comparison is conducted for both total trust and each category of trust perception (performance, process, and purpose) across all scenarios together as well as within each of the three individual scenarios: the performance scenario (S1), the process scenario (S2), and the purpose scenario (S3).\\
Figure \ref{fig:total_trust} displays the updated total trust for the positive update groups compared to the negative update groups, revealing a noticeable difference. To further assess this difference, we conducted a series of two-tailed and one-tailed t-tests, comparing the variations between the positive and negative update groups for updated total trust, as well as the performance, process, and purpose related trust perceptions reported by participants within each scenario and when considering all scenarios together.
In these t-tests, the null hypothesis assesses that the samples are generated from distributions with the same mean, while the alternative hypothesis suggests that they are derived from different distributions. Our results from the t-tests support H1, indicating that the updated total trust, as well as the updated performance, process, and purpose related  perceptions, significantly differ in the positive update groups compared to the negative update groups. This statistical significance holds true for all scenarios collectively (all p-values $< 0.0001(^{****})$), the performance scenario (S1) (all p-values $< 0.0001(^{****})$), and the purpose scenario (S3) (all p-values $< 0.01$). However, in the process scenario (S2), only the process perception of trust demonstrates statistically significant differences between the positive and negative update groups ($t(32.64) = -2.42$, $p = 0.01$).

\paragraph{H2- Positive update groups induce increase in trust}
For hypothesis H2, we examine whether the positive update groups exhibit an increase in trust compared to their initial trust. This examination is conducted for total trust and each information of trust (performance, process, and purpose) across all scenarios together and within each of the three individual scenarios: the performance scenario (S1), the process scenario (S2), and the purpose scenario (S3). \\
Figure \ref{fig:total_trust} illustrates that total trust increased in the updated state when compared to the initial state for the positive update groups. We conducted one-tailed t-tests to determine whether total trust and each information of trust exhibited a statistically significant increase from the initial state to the updated state within each scenario and across all scenarios together.\\
The results of these t-tests\footnote{complete details in supplementary material} support H2. The data show that total trust, as well as all perceptions of trust, was significantly different from the initial state to the updated state, and descriptively higher (all p-values $< 0.05$). 

\paragraph{H3- Negative update groups induce decrease in trust}

To test hypothesis H3, we compare the updated trust and initial trust within the negative update groups for total trust and each category of trust information (performance, process, and purpose) across all scenarios collectively, and within each of the three individual scenarios: the performance scenario (S1), the process scenario (S2), and the purpose scenario (S3).\\
As demonstrated in Figure \ref{fig:total_trust}, total trust decreased in the updated state when compared to the initial state for the negative update groups in the combined scenarios and in each scenario except for the process scenario (S2). In the case of the negative update within the process scenario (S2), the robot achieved the goal and could perform the task but not in the desired and preferred human process. This suggests that the effect of  process is not as strong on trust when other factors are satisfied (i.e., performance), and even a negative update for the process can increase trust.\\
To test the statistical significance of our observations, we conducted a series of one-tailed t-tests to determine whether total trust and each information of trust exhibited a significant decrease from the initial state to the updated state within each scenario and across all scenarios collectively.\\
The results reveal that total trust and all information categories of trust decreased from the initial state to the updated state with statistically significant p-values in the performance scenario (S1) (p-values $< 0.01$). In the collective scenarios, both total trust and purpose perception of trust were significantly different from the initial state to the updated state, and descriptively lower ($t(118) = 1.6$, $p = 0.05$ and $t(118) = 1.7$, $p= 0.046$). However, while the performance and process perceptions of trust also decreased in the collective scenarios, they were not significantly different. We attribute this in part to the inclusion of data from the process scenario in this collection, which exhibited an increase in trust.\\
Regarding the purpose scenario (S3), the total trust and all perceptions of trust decreased from the initial state to the updated state, but these decreases were not statistically significant. Conversely, in the process scenario (S2), total trust and each perception of trust increased in the updated state, with statistical significance for all perceptions (p-values $< 0.05$) except the process perception of trust. This highlights the influence of controlling the process information of the model on the process perception of trust.

\paragraph{H4- Correlation of performance scenarios (S1) with performance information of Trust}
In hypothesis H4, we aim to determine whether the increase or decrease in the performance perception of trust in the performance scenario (S1) is more significant than other perceptions, i.e. process and purpose, in the positive or negative update groups, respectively.\\
Comparing Cohen's-d effect size we observe that the increase in performance perception of trust in the positive update groups ($1.26> 0.59, 0.9$) and the decrease in performance perception of trust in the negative update groups ($1.83 > 0.7, 1.31$) have the highest effect size when compared to the changes in process and purpose perception of trust. Moreover, the results from a one-way ANOVA test demonstrate that the increase and decrease in these perceptions of trust are significantly different from each other in the positive ($F(2, 60) = 3.75$, $p = 0.03$) and negative update groups ($F(2, 60) = 5.00$, $p = 0.009$) respectively.

\paragraph{H5 Correlation of process scenarios (S2) with process information of Trust}
To validate hypothesis H5, we examine whether, in the process scenario (S2), the increase or decrease in process perception of trust is more significant than performance and purpose perception of trust within positive or negative update groups.
Comparing Cohen's-d effect size, we find that in the positive update groups, the effect size for the increase in process perception of trust is more than the purpose perception ($0.88 > 0.68$), although it is more in the performance perception ($1.09$). This result is not surprising, as a successful process typically accompanies successful performance. In the negative update groups, we observe that trust increases in all perceptions; however, the increase in process perception of trust has the smallest effect size compared to trust in performance and purpose ($0.1 < 1.21, 0.53$). This suggests that while negative updates to the process does not decrease overall trust, it has a relatively weaker positive impact on the perception of the process compared to other perceptions. While the results from a one-way ANOVA test in the positive update groups do not show significant differences in the increase of various perceptions ($F(2, 63) = 0.46$, $p = 0.63$), in the negative update groups, the decrease in these perceptions of trust is  significantly different ($F(2, 60) = 4.77$, $p = 0.01$).

\paragraph{H6- Correlation of purpose scenarios (S3) with purpose information of Trust}
Here, we evaluate if in the purpose scenario (S3), the increase or decrease of purpose perception of trust with positive or negative updates is the most significant one among other perceptions. By comparing the Cohen's-d effect size, we can see that in the positive update group, the increase in purpose perception of trust has greater effect size than the process perception ($0.56>0.54$), although it is still lower than the performance perception ($0.69$). Similar to the process scenario, this result is not surprising because a successful purpose typically involves a successful performance, and it is challenging to disentangle the two. However, in the negative update group, we can observe that the purpose perception of trust decreases with higher effect size than the performance and process perceptions ($0.27 > 0.08, 0.046$), even though the results from a one-way ANOVA test do not show significant differences between the increase and decrease in various perceptions of trust for both the positive ($F(2, 57) = 0.18$, $p = 0.83$) and the negative groups ($F(2, 51) = 0.27$, $p = 0.76$) respectively.

\subsection{Discussion}
Our results show that change in likelihood of the model does result in change in human trust, as assessed by trust questionnaire. Furthermore, our findings reveal that both reductions and increases in likelihood have varying effects on trust. Notably, we observed a significant increase in trust across all three trust perception categories (performance, process, and purpose) with a positive update of the likelihood of contract fulfillment. On the other hand, decreasing the likelihood of specific model information, while causing a decrease in trust, did not uniformly decrease trust across all trust perception categories. This seems to support other research that indicates trust and distrust may need to be studied separately rather than as opposing ends of the same linear scale \cite{Schroeder_Chiou_Craig_2021}. 

In the Performance Scenario (S1), we observed a significant decrease in trust across all trust perceptions when we reduced the likelihood of contract achievement, focusing on performance aspect of the model. However, for the other scenarios, namely process (S2) and purpose (S3), reducing the likelihood of specific information did not significantly decrease trust across all trust perceptions. This disparity can be attributed to the fact that reducing the likelihood of achieving the contract for information related to process or purpose does not necessarily entail a consistent or reduction in performance-related information. For example, in the negative process scenario (S2), while the likelihood of achieving the contract with a specific process decreased, the likelihood of the robot's capability to perform the task and the likelihood of the robot adhering to the user's intended purpose increased. Consequently, this delicate interplay led to an increase in trust even with a negative process information update. 

Furthermore, the significant of results for performance perception of trust in the performance scenario and other scenarios, emphasizes the substantial impact of robot performance in shaping trust in comparison to other aspects. However, an alternative interpretation of this impact could be attributed to the assumption that performance naturally includes an aligned purpose and a certain level of satisfactory process. In other words, we cannot assume that performance is achieved without ensuring the robot's purpose aligns with the user's purpose, just as we cannot assume that performance is attained without meeting a degree of effective and ethical processes.

Moreover, though we aimed to control for the other trust perceptions in each scenario, our findings emphasize the high degree of correlation between the three trust information categories in the model. This signifies the inherent challenge of designing experiments that can simulate practical circumstances while completely isolating one trust information category from the others.

Another noteworthy observation is that updates to the human model are consistently associated with the human acquiring more information about the robot's processes. Consequently, both positive and negative updates tend to enhance clarity in certain aspects of the process. This makes it challenging to distinguish between the clarity of the overall process, which can also convey a negative view of the robot, and the lack of a clear understanding of the robot's process when implementing model updates. This emphasizes the need for a modified questionnaire that can effectively differentiate between various aspects of the process-related information.

In conclusion, our results shed light on the dynamics of trust, trust information categories, and model updates. While the results did not confirm all aspects of the hypotheses with statistically significant results, the insights derived from our results align with our mental model-based framework of trust, showing that our framework can effectively capture the aforementioned dynamics.

\section{Conclusion}
In this paper, we formalized a mental model based theory of trust that can be used as a basis to infer human trust. Using our proposed framework, we further formalized the notions of human reliance, appropriate trust, and provide mechanisms to model trust evolution; all of which can be used as a foundation for any future trust-aware decision-making frameworks. We further ran human subject studies to investigate if adjusting individuals' beliefs about the agent, as outlined in our framework, induces changes in trust perception, and if modifying a specific model component brings about a corresponding change in trust perceptions related to performance, process, and purpose information. In future work, we hope to develop trust-aware interpretable behavior generation and explanation generation methods that leverages our framework of trust. We also plan to develop decision-making frameworks that can leverage our mental model based theory of trust to guide longitudinal human-robot interactions.
\section*{Acknowledgments}
This research is supported in part by ONR grants N00014-18-1-2442, N14-18-1-2840 and N00014-23-1-2409 and a JP Morgan AI Faculty Research grant to Kambhampati.

\bibliographystyle{unsrt}  
\bibliography{main}  

\appendix
\section*{Appendix A -- Trust Questionnaire}
As it mentioned in the paper, we employ a trust questionnaire developed by Chancey et al.. Since the original questionnaire was crafted within the context of an operator and a recommendation system, specifically, a tank-spotting aid, we have carefully revised the questions to adapt the questionnaire for use in our domain. We ensure that they capture the core essence of the original questionnaire. The questionnaire we used in the study is given in Table \ref{tab:trust-q}.
\begin{table}[th]
    \centering
    \caption{Trust Questionnaire}
    \resizebox{\textwidth}{!}{\begin{tabular}{p{18cm}l}
    \toprule
    \textbf{Performance (Predictability; Ability): What Does the Automation Do?}\\
    ~~~~~~The robot always completes the task that is required of it.\\
    ~~~~~~The robot reliably completes the task that is required of it.\\
    ~~~~~~The robot consistently performs the task that is required of it.\\
    ~~~~~~I can rely on the robot to function properly.\\
    ~~~~~~The robot adequately completes the block-stacking task that is required of it.\\
    \textbf{Process (Dependability; Integrity): How Does the Automation Work?}\\
    ~~~~~~Although I might not fully understand the robot's internal processes, I can predict how it completes a task.\\
    ~~~~~~I will be able to predict how the robot will perform in the future.\\
    ~~~~~~I understand why the robot completed the task in that manner.\\
    ~~~~~~It is easy to follow why the robot completed the task in that manner.\\
    ~~~~~~I recognize how to assess whether the robot is completing its task well.\\ 
    \textbf{Purpose (Faith; Benevolence): Why Was the Automation Developed?}\\
    ~~~~~~I believe the robot will be able to complete the task required of it, even when I don't know for certain that the robot can do it.\\
    ~~~~~~When I am uncertain about deciding whether the robot will complete the task, I believe the robot will do it.\\
    ~~~~~~When I am not sure about whether the robot will complete the task, I have faith that the robot will complete the task.\\
    ~~~~~~Even when the robot completes the task in an unusual way, I am certain that the robot will complete its task.\\
    ~~~~~~Even if I have no reason to expect that the robot will function properly, I still feel certain that it will complete the tasks required of it.\\
        \bottomrule
    \end{tabular}}
    \label{tab:trust-q}
\end{table}
\section*{Appendix B -- Human Subjects Studies Procedure}
\begin{figure*}[h]
    \centering
    \includegraphics[width=\textwidth]{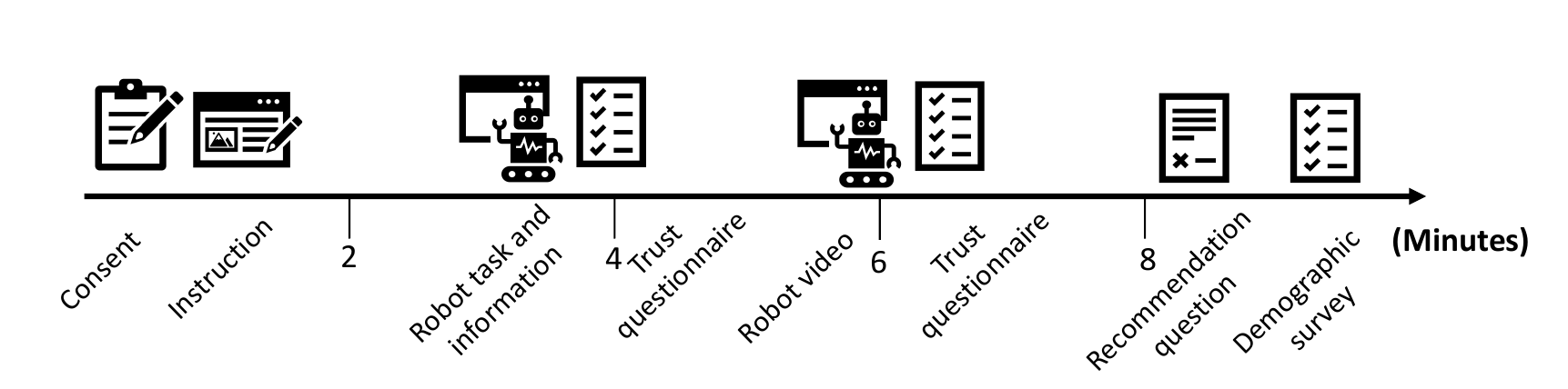}
    \caption[Study Procedure Overview.]{Study procedure overview.}
    \label{fig:procedure}
\end{figure*}

As depicted in Figure \ref{fig:procedure}, upon granting their consent, each participant is directed to the instruction page. Within these instructions, we introduce a gamified element to the study. Participants assume the role of technology purchasers tasked with making a critical recommendation to the CEO of their company regarding the purchase of a robot. The success of their company's future depends on their final decision. A mistake in their recommendation can reflect poorly for their standing as employees within the company. We also provide a brief overview of the study's process.
Subsequently, participants are presented with a description of the task that the robot is expected to perform, accompanied by an image depicting the robot within its task environment. This image includes annotated objects for clarity. Furthermore, participants receive information about the robot and the setting, aligning with the scenarios outlined in the study. Following this phase, participants are asked to complete a fifteen-item trust questionnaire, which is accompanied by two attention check questions, all presented in random order.\\
Next, participants view a video showcasing the robot's execution of the task, after which they are prompted to respond to a similar set of questions. This phase requires them to consider the observations made during the video presentation.\\
Subsequently, participants are invited to provide their recommendation regarding whether their company should proceed with the purchase of the robot or not.\\
Lastly, participants are asked to complete a set of demographic questions.

\section*{Appendix C -- Additional Results}
\begin{figure*}[h]
\begin{subfigure}[\label{fig:perf_pos}]
    \centering
    \includegraphics[width=0.45\textwidth]{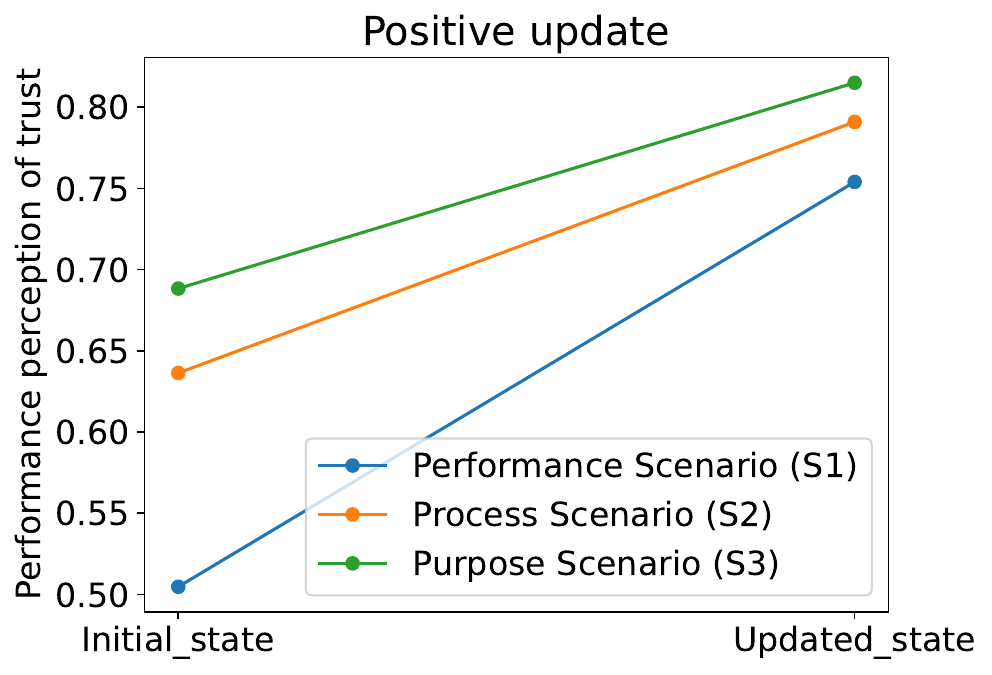}
    \hfil
    \end{subfigure}
    \begin{subfigure}[\label{fig:perf_neg}]
    \centering
    \includegraphics[width=0.45\textwidth]{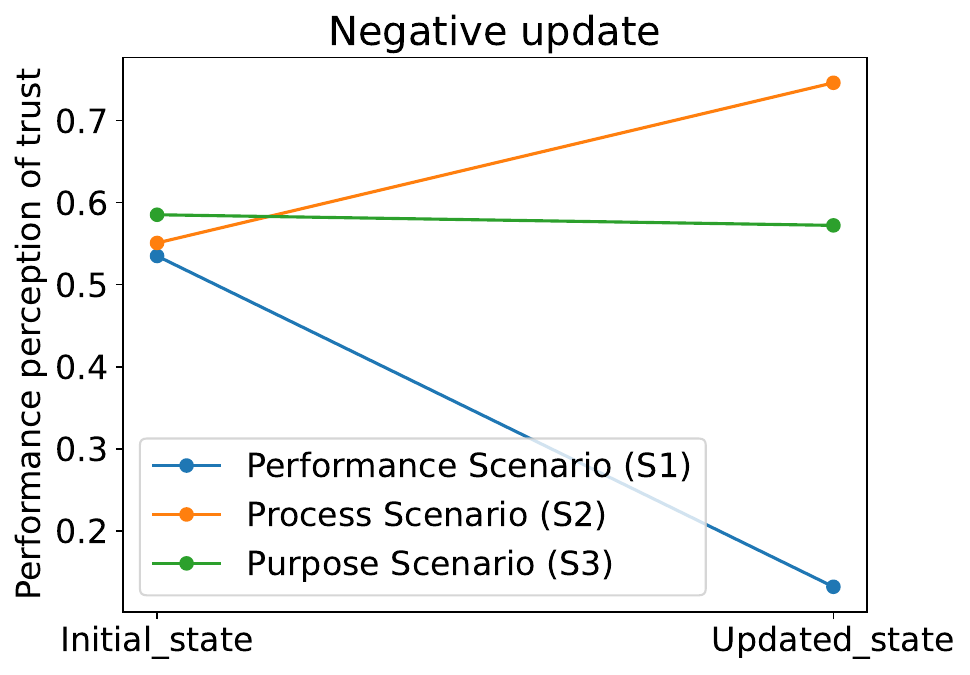}
    \hfil
    \end{subfigure}
    \caption[The Change in Performance Perception of Trust Across Different Scenarios With (a) Positive Updates and (b) Negative Updates.]{The change in performance perception of trust across different scenarios with (a) positive updates and (b) negative updates.}
    \label{fig:Perf}  
\end{figure*}

Figure \ref{fig:perf_pos} and \ref{fig:perf_neg} provide visual evidence supporting that the performance perception has increased or decreased the most in the performance scenario (S1) compared to other scenarios for positive and negative update groups, respectively.

\begin{figure*}[h]
\begin{subfigure}[\label{fig:proc_pos}]
    \centering
    \includegraphics[width=0.45\textwidth]{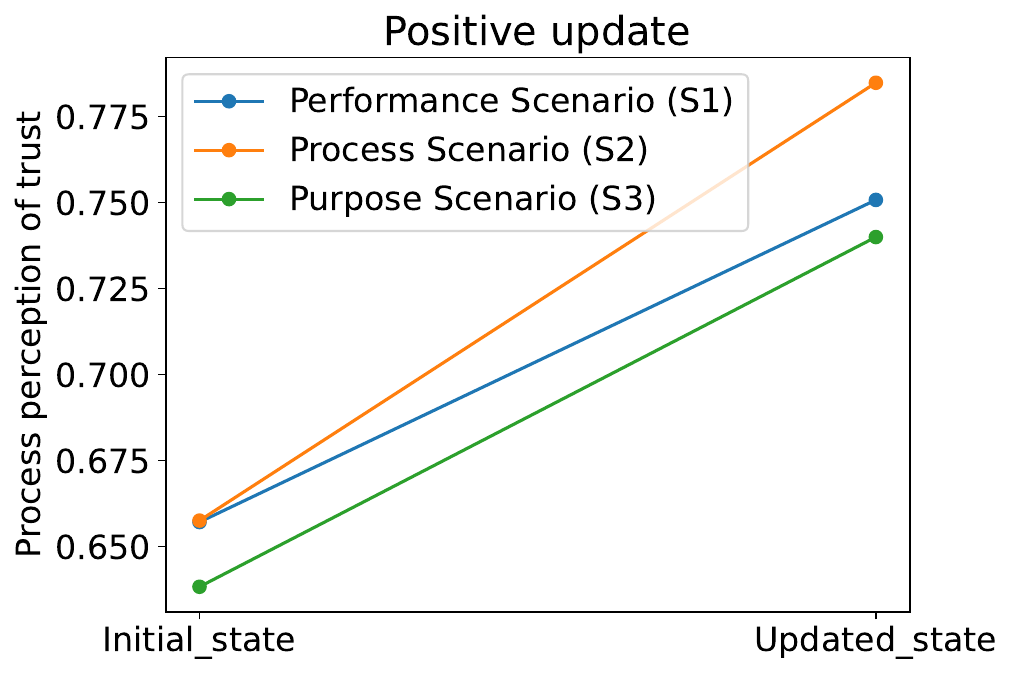}
    \hfil
    \end{subfigure}
    \begin{subfigure}[\label{fig:proc_neg}]
    \centering
    \includegraphics[width=0.45\textwidth]{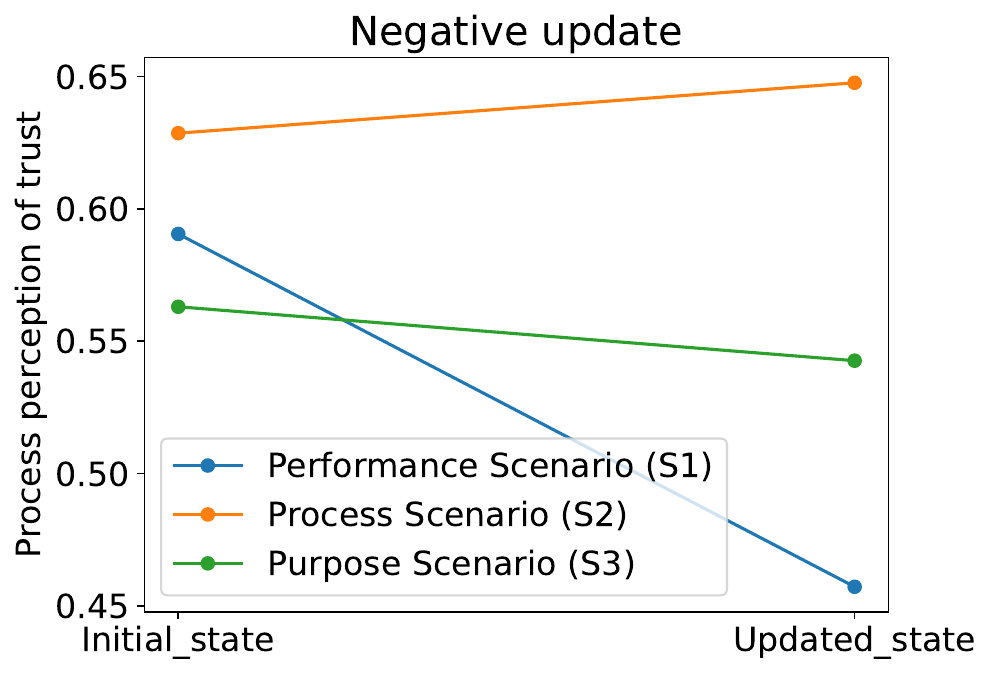}
    \hfil
    \end{subfigure}
    \caption[The Change in Process Perception of Trust Across Different Scenarios With (a) Positive Updates and (b) Negative Updates.]{The change in process perception of trust across different scenarios with (a) positive updates and (b) negative updates.}
    \label{fig:Proc}  
\end{figure*}

Furthermore, we can gain further insights into how different scenarios affect process perception of trust by examining Figure \ref{fig:proc_pos} and \ref{fig:proc_neg}. Figure \ref{fig:proc_pos} demonstrates that the process perception has increased the most in the process scenario (S2). However, according to Figure \ref{fig:proc_neg}, with negative updates, the process perception of trust decreases the most in the performance scenario (S1) rather than in the process scenario (S2).

\begin{figure*}[h]
\begin{subfigure}[\label{fig:pur_pos}]
    \centering
    \includegraphics[width=0.45\textwidth]{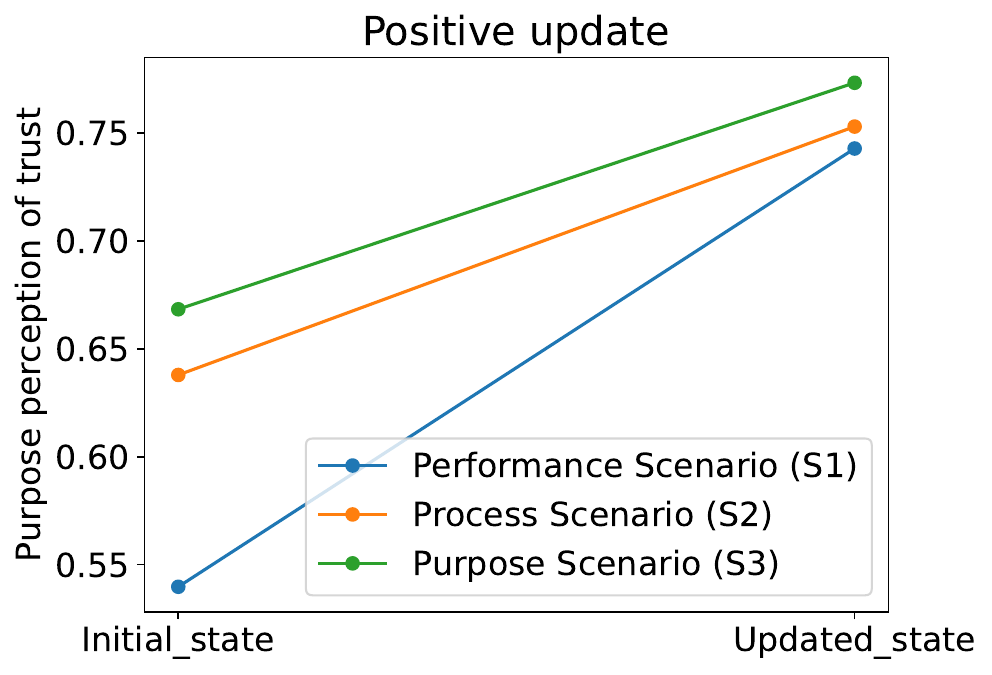}
    \hfil
    \end{subfigure}
    \begin{subfigure}[\label{fig:pur_neg}]
    \centering
    \includegraphics[width=0.45\textwidth]{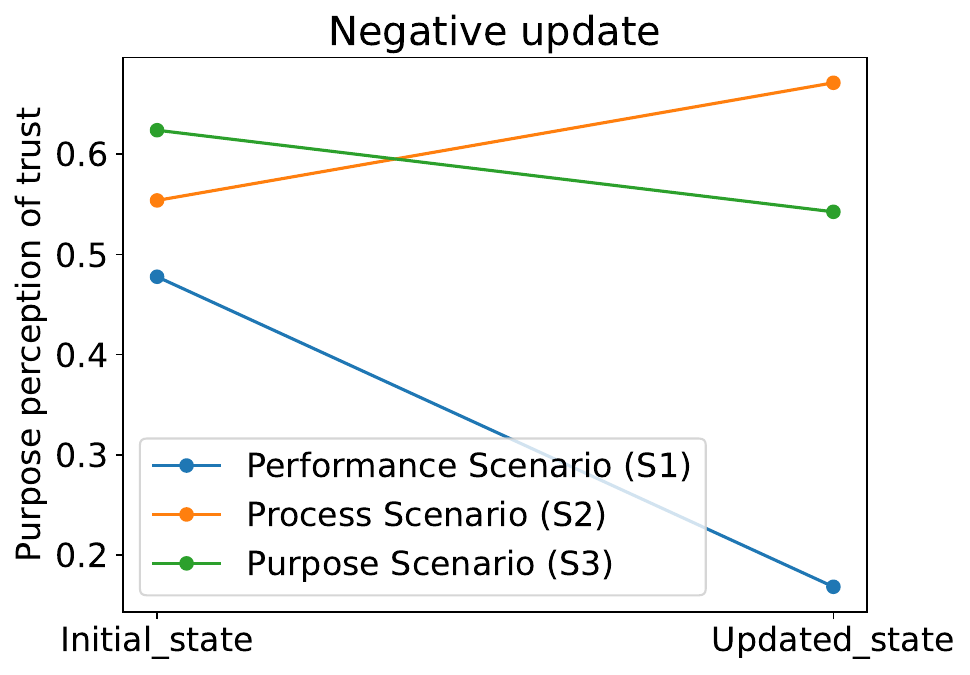}
    \hfil
    \end{subfigure}
    \caption[The Change in Purpose Perception of Trust Across Different Scenarios With (a) Positive Updates and (b) Negative Updates.]{The change in purpose perception of trust across different scenarios with (a) positive updates and (b) negative updates.}
    \label{fig:Purp}  
\end{figure*}

We also assess if the purpose scenario (S3) results in the most increase or decrease in purpose perception when compared to other scenarios. Figures \ref{fig:pur_pos} and \ref{fig:pur_neg} demonstrate that the performance scenario (S1) leads to more significant increases or decreases in purpose perception of trust than the purpose scenario (S3) with both positive and negative updates.

\section*{Appendix D -- Statistical Tests}
We performed t-tests to test each hypothesis in the paper for each scenario as well as across all scenarios. We utilized a two-tailed t-test for the first hypothesis and a one-tailed t-test for the subsequent hypotheses. Any claims of statistical significance will be made against a significance level of $0.05$. The complete detail of the data from the statistical tests are in Tables \ref{tab:S1}, \ref{tab:S2}, \ref{tab:S3}, and \ref{tab:all}.
\begin{table}[t]
\centering
\caption[T-Test Results for Performance Scenario (S1): (a) Trust Comparison Between Positive and Negative Update Groups in Updated State, (b) Trust Increase in Positive Update Groups, and (c) Trust Decrease in Negative Update Groups.]{T-Test Results for Performance Scenario (S1): (a) Trust comparison between positive and negative update groups in updated state, (b) Trust increase in positive update groups, and (c) Trust decrease in negative update groups.}
\label{tab:S1}
\resizebox{.8\textwidth}{!}{\begin{tabular}{clccccc}
\multicolumn{7}{c}{Performance Scenario (S1)}\\
\cmidrule{1-7}
\multicolumn{2}{c}{Trust Perception}& \multicolumn{5}{c}{T-tests}\\
\cmidrule{5-6}
&& T & dof & p-val. & cohen-d & tail\\
\cmidrule{1-7}
\shortstack{(a) Positive vs. Negative \\ \small{updated\_state}}\\
\midrule
Total trust &&$-11.03$ & $40$ & $< 0.0001(^{****})$ & $3.404$ & two-sided\\
Performance inf. && $-10.98$ & $40$ & $< 0.0001(^{****})$ & $3.39$ & less \\
Process inf. && $-5.07$ & $40$ & $< 0.0001(^{****})$ & $1.57$ & less\\
Purpose inf. && $-10.45$ & $40$ & $< 0.0001(^{****})$ & $3.22$ & less\\
\midrule
\shortstack{(b) Positive update groups \\ \small{updated\_state > initial\_state}}\\
\midrule
Total trust &&$-3.69$ & $40$ & $0.0003$ & $1.14$ & less\\
Performance inf. &&$-4.059$ & $40$ & $0.00019$ & $1.259$& less\\
Process inf. & &$-1.90$ & $40$ & $0.03$ & $0.59$ & less \\
Purpose inf. &&$-2.91$ & $40$ & $0.003$ & $0.90$ & less\\
\midrule
\shortstack{(c) Negative update groups \\ \small{updated\_state < initial\_state}}\\
\midrule
Total trust && $4.75$ & $40$ & $0.00001$ & $1.47$& greater\\
Performance inf. & &$5.949$ & $40$ & $< 0.0001(^{****})$ & $1.83$ & greater\\
Process inf. &&$2.28$ & $40$ & $0.01$ & $0.70$ & greater \\
Purpose inf. &&$4.24$ & $40$ & $< 0.0001(^{****})$ & $1.31$ & greater \\
\bottomrule
\end{tabular}}
\end{table}

\begin{table}[t]
\centering
\caption[T-Test Results for Process Scenario (S2): (a) Trust Comparison Between Positive and Negative Update Groups in the Updated State, (b) Trust Increase in the Positive Update Groups, and (c) Trust Increase in the Negative Update Groups.]{T-Test Results for Process Scenario (S2): (a) Trust comparison between positive and negative update groups in the updated state, (b) Trust increase in the positive update groups, and (c) Trust increase in the negative update groups.}
\label{tab:S2}
\resizebox{.8\textwidth}{!}{\begin{tabular}{clccccc}
\multicolumn{7}{c}{Process Scenario (S2)}\\
\cmidrule{1-7}
\multicolumn{2}{c}{Trust Perception}& \multicolumn{5}{c}{T-tests}\\
\cmidrule{5-6}
&& T & dof & p-val. & cohen-d & tail\\
\cmidrule{1-7}
\shortstack{(a) Positive vs. Negative \\ \small{updated\_state}}\\
\midrule
Total trust &&$-1.71$ & $34.53$ & $0.096$ & $0.53$ & two-sided \\
Performance inf. && $-0.91$ & $38.52$ & $0.18$ & $0.28$ & less\\
Process inf. && $-2.42$ & $32.64$ & $0.01$ & $0.75$ & less \\
Purpose inf. && $-1.27$ & $32.76$ & $0.11$ & $0.39$ & less \\
\midrule
\shortstack{(b) Positive update groups \\ \small{updated\_state > initial\_state}}\\
\midrule
Total trust &&$-3.22$ & $42$ & $0.001$ & $0.97$ & less \\
Performance inf. & &$-3.61$ & $42$ & $0.0004$ & $1.09$ & less\\
Process inf. & &$-2.92$ & $42$ & $0.003$ & $0.88$ & less \\
Purpose inf. & &$-2.24$ & $42$ & $0.015$ & $0.68$ & less \\
\midrule
\shortstack{(c) Negative update groups \\ \small{updated\_state > initial\_state}}\\
\midrule
Total trust && $-2.11$ & $40$ & $0.02$ & $0.65$ & less\\
Performance inf. & & $-3.92$& $40$&$0.0002$&$1.21$&less\\
Process inf. & &$-0.32$ & $40$ & $0.37$ & $0.10$ & less \\
Purpose inf. &&$-1.68$ & $40$ & $0.05$ & $0.53$ & less \\
\bottomrule
\end{tabular}}
\end{table}

\begin{table}[t]
\centering
\caption[T-Test Results for Purpose Scenario (S3): (a) Trust Comparison Between Positive and Negative Update Groups in the Updated State, (b) Trust Increase in the Positive Update Groups, and (c) Trust Decrease in the Negative Update Groups.]{T-Test Results for Purpose Scenario (S3): (a) Trust comparison between positive and negative update groups in the updated state, (b) Trust increase in the positive update groups, and (c) Trust decrease in the negative update groups.}
\label{tab:S3}
\resizebox{.8\textwidth}{!}{\begin{tabular}{clccccc}
\multicolumn{7}{c}{Purpose Scenario (S3)}\\
\cmidrule{1-7}
\multicolumn{2}{c}{Trust Perception}& \multicolumn{5}{c}{T-tests}\\
\cmidrule{5-6}
&& T & dof & p-val. & cohen-d & tail\\
\cmidrule{1-7}
\shortstack{(a) Positive vs. Negative \\ \small{updated\_state}}\\
\midrule
Total trust &&$-2.80$ & $24.07$ & $0.01$ & $0.94$ & two-sided\\
Performance inf. && $-2.78$ & $22.04$ & $0.005$ & $0.94$ & less\\
Process inf. && $-2.31$ & $26.84$ & $0.01$ & $0.77$ & less\\
Purpose inf. && $-2.61$ & $25.25$ & $0.007$ & $0.87$ & less\\
\midrule
\shortstack{(b) Positive update groups \\ \small{updated\_state > initial\_state}}\\
\midrule
Total trust &&$-2.10$ & $38$ & $0.02$ & $0.66$ & less\\
Performance inf. &&$-2.19$ & $38$ & $0.02$ & $0.69$ & less\\
Process inf. & &$-1.70$ & $38$ & $0.048$ & $0.54$ & less \\
Purpose inf. & &$-1.78$ & $38$ & $0.04$ & $0.56$ & less\\
\midrule
\shortstack{(c) Negative update groups \\ \small{updated\_state < initial\_state}}\\
\midrule
Total trust && $0.44$ & $34$ & $0.33$ & $0.15$ & greater\\
Performance inf. &&$0.14$ & $34$ & $0.44$ & $0.046$ & greater\\
Process inf. & &$0.23$ & $34$ & $0.41$ & $0.08$ & greater \\
Purpose inf. &&$0.82$ & $34$ & $0.21$ & $0.27$ & greater\\
\bottomrule
\end{tabular}}
\end{table}

\begin{table}[t]
\centering
\caption{T-Test Results for Data Collected Across All Scenarios Together.}
\label{tab:all}
\resizebox{.8\textwidth}{!}{\begin{tabular}{clccccc}
\multicolumn{7}{c}{All Scenarios}\\
\cmidrule{1-7}
\multicolumn{2}{c}{Trust Perception}& \multicolumn{5}{c}{T-tests}\\
\cmidrule{5-6}
&& T & dof & p-val. & cohen-d & tail\\
\cmidrule{1-7}
\shortstack{(a) Positive vs. Negative \\ \small{updated\_state}}\\
\midrule
Total trust &&$-6.59$ & $81.81$ & $< 0.0001(^{****})$ & $1.21$ & two-sided\\
Performance inf. & &$-6.13$ & $79.47$ & $< 0.0001(^{****})$ & $1.12$ & less\\
Process inf. && $-5.37$ & $97.69$ & $< 0.0001(^{****})$ & $0.98$ & less\\
Purpose inf. && $-6.20$ & $83.37$ & $< 0.0001(^{****})$ & $1.14$ & less\\
\midrule
\shortstack{(b) Positive update groups \\ \small{updated\_state > initial\_state}}\\
\midrule
Total trust &&$-5.14$ & $124$ & $< 0.0001(^{****})$ & $0.92$ & less\\
Performance inf. & &$-5.46$ & $124$ & $< 0.0001(^{****})$ & $0.97$ & less\\
Process inf. & &$-3.73$ & $124$ & $0.0001$ & $0.66$ & less\\
Purpose inf. & &$-4.04$ & $124$ & $< 0.0001(^{****})$ & $0.72$ & less\\
\midrule
\shortstack{(c) Negative update groups \\ \small{updated\_state < initial\_state}}\\
\midrule
Total trust && $1.60$ & $118$ & $0.05$ & $0.29$ & greater\\
Performance inf. &&$1.45$ & $118$ & $0.07$ & $0.26$ & greater\\
Process inf. & &$1.15$ & $118$ & $0.13$ & $0.21$ & greater\\
Purpose inf. &&$1.70$ & $118$ & $0.046$ & $0.31$ & greater\\
\bottomrule
\end{tabular}}
\end{table}
\end{document}